\documentclass[12pt]{article}
\usepackage{amsfonts}
\usepackage{a4wide}
\usepackage{amsmath,amscd,amsthm,a4,amssymb}

\begin{document}

\begin{center}
{\Large \textbf{Addressing common misinterpretations of KART and UAT in
neural network literature}}

\

\textsc{Vugar E. Ismailov} \

\bigskip

Institute of Mathematics and Mechanics, Baku, Azerbaijan

Center for Mathematics and its Applications, Khazar University, Baku,
Azerbaijan

{e-mail:} {vugaris@mail.ru}
\end{center}

\smallskip

\textbf{Abstract.} This note addresses the Kolmogorov-Arnold Representation
Theorem (KART) and the Universal Approximation Theorem (UAT), focusing on
their frequent misinterpretations found in the neural network literature.
Our remarks aim to support a more accurate understanding of KART and UAT
among neural network specialists. In addition, we explore the minimal number
of neurons required for universal approximation, showing that the same
number of neurons needed for exact representation of functions in KART-based
networks also suffices for standard multilayer perceptrons in the context of
approximation.

\bigskip



\bigskip

\begin{center}
{\large \textbf{Introduction}}
\end{center}

The increasing integration of neural networks into various fields of science
and engineering has sparked significant interest in the theoretical
foundations of approximation by these networks. Two central results in this
context are the Kolmogorov-Arnold Representation Theorem (KART) and the
Universal Approximation Theorem (UAT), both of which provide essential
insights into the capacity of neural networks to approximate complicated
multivariate functions. However, numerous works in the literature exhibit
misunderstandings concerning the scope and limitations of these theorems.

In recent years, KART has gained considerable attention following the
emergence of Kolmogorov-Arnold Networks (KAN), which are based on the
principles established in KART. Despite their practical successes, the full
implications of KART are often misunderstood or misrepresented in the
literature. One major misconception is that the original KART applies to
continuous functions defined on arbitrary bounded sets of the $d$%
-dimensional Euclidean space, or even to all multivariate functions. Another
common misunderstanding is that UAT always requires increasing the number of
neurons to achieve greater approximation accuracy.

In this paper, we aim to address these misconceptions, presenting a
generalization of KART that extends its applicability to discontinuous
functions on the unit cube, as well as to multivariate functions (both
continuous and discontinuous) defined on bounded sets beyond the unit cube.
Although many works in the literature treat such extensions as if they
follow directly from the original KART, this is not the case. The original
theorem applies only to continuous functions on the unit cube or, more
generally, on compact subsets of $\mathbb{R}^{d}$. The broader claims found
in the literature are valid only in light of the generalization we
established in our recent works --- yet these claims are often incorrectly
attributed to the original KART, rather than to the extended framework
introduced in \cite{Is1} and \cite{Isma}.

Furthermore, we challenge the prevailing notion that the number of hidden
neurons required in neural networks always depends on the approximation
error. In contrast to this belief, we show that for certain types of
activation functions, a fixed number of neurons can achieve arbitrary
approximation accuracy for all continuous functions.

The question of the minimum number of neurons required for universal
approximation is also addressed in this paper. We show that the bounds on
number of neurons required for exact representation of functions in
KART-based networks also apply to standard multilayer perceptrons (MLPs)
with smooth activation functions in the context of approximation, thereby
providing a clearer picture of the relationship between KART-based networks
and MLPs. While these bounds are known to be minimal in KART-based networks,
we conjecture that they are also minimal for MLPs.

In summary, this paper seeks to provide a clearer and more accurate
understanding of KART, UAT, and their applications in neural network theory.
By addressing critical misconceptions and offering new theoretical insights,
we aim to support more rigorous and informed use of these foundational
results in future studies.

\medskip

\noindent\textbf{Our contributions.} The main contributions of this paper are:
\begin{itemize}
  \item A rigorous generalization of KART, showing that its representation power
  extends to all multivariate functions on bounded domains, including
  discontinuous ones (Theorem~1 and Corollary~1).
  \item Clarification of the dimensional limitations of UAT for single-hidden-layer
  networks, emphasizing that fixed-width universality holds only in the univariate
  case (Theorems~2, 3).
  \item Proving that single-hidden-layer networks with just one hidden neuron can approximate
  any continuous univariate function (Theorem~4).
  \item A new universality result for deep networks with fixed weights,
  establishing an explicit hidden neuron bound of $3d+1$, sufficient for
  universal approximation (Theorem~6).
  \item A conjecture on the minimal neuron count required for universality in multivariate settings.
\end{itemize}
We stress again that these results are theoretical in nature, intending to
clarify widespread misconceptions concerning KART and UAT, while establishing broader domains for
KART-based networks and providing explicit neuron-count bounds for MLPs that
may guide future applied research.

\bigskip

\bigskip

\bigskip

\begin{center}
{\large \textbf{Remark 1: Kolmogorov-Arnold representation theorem}}
\end{center}

The Kolmogorov-Arnold Representation Theorem (KART) is important for
understanding the theoretical potential of neural networks. It states that
for the unit cube $\mathbb{I}^{n},~\mathbb{I}=[0,1],~n\geq 2,$ there exist $%
n(2n+1)$ universal, continuous, one-variable functions $\varphi
_{q,p}(x_{p}),$ $q=1,...,2n+1,$ $p=1,...,n,$ such that each function $f\in C(%
\mathbb{I}^{n})$ admits the precise representation
\begin{equation*}
f(x_{1},...,x_{n})=\sum_{q=1}^{2n+1}\Phi _{q}(\sum_{p=1}^{n}\varphi
_{q,p}(x_{p})),\eqno(1)
\end{equation*}%
where $\Phi _{q}$ are continuous one-variable functions depending on $f$
(see \cite{Kol}).

Formula (1) describes a feedforward neural network with the following
structure:

\begin{itemize}
\item \textbf{Input Layer:} This layer has $n$ neurons, which receive input
signals $x_{1},...x_{n}$;

\item \textbf{First Hidden Layer:} This layer consists of $n(2n+1)$ neurons.
The $(q,p)$-th neuron $y_{q,p}$ ($1\leq q\leq 2n+1,$ $1\leq p\leq n$)
produces an output $\varphi _{q,p}(x_{p})$, where $\varphi _{q,p}$
represents the activation function applied to the input $x_{p}$;

\item \textbf{Second Hidden Layer:} This layer consists of $2n+1$ neurons.
Each neuron $z_{q}$ ($1\leq q\leq 2n+1$) in this layer produces an output $%
\Phi _{q}\left( \sum_{p=1}^{n}y_{q,p}\right) $, where $\Phi _{q}$ represents
the activation function applied to the sum of certain outputs from the
previous layer;

\item \textbf{Output Layer:} This layer has a single neuron that sums up the
outputs from the second hidden layer to produce the final output $%
f(x_{1},...x_{n})$.
\end{itemize}

Thus, KART shows that every continuous multivariate function can be
implemented by a feedforward neural network. It is important to address a
common misinterpretation of KART found in various articles. For instance,
the recent influential paper \cite{Liu}\ introducing Kolmogorov-Arnold
Networks (KANs), which are inspired by KART, states: \textquotedblleft In a
sense, KART showed that the only true multivariate function is addition,
since every other function can be written using univariate functions and
sum". This or similar misinterpretations are also evident in other
literature related to KANs. For example, the following sentences illustrate
this misunderstanding:

(a) \textquotedblleft This theorem tells us that any multivariate function
can essentially be decomposed into the sum of functions of sums" \cite{Boz}.

(b) \textquotedblleft This formulation demonstrates that multivariate
functions can fundamentally be reduced to a suitably defined composition of
univariate functions, where the composition only involves simple addition"
\cite{Kia}.

(c) \textquotedblleft This implies that addition is the only true
multivariate operation, while other multivariate operations (including
multiplication) can be expressed as additions combined with univariate
functions" \cite{Liu2}.

There are many other papers erroneously formulating KART for arbitrary
multivariate functions (see, e.g., \cite{Bresson,Carlo,Peng}).

However, it is essential to note that while the \emph{original KART} guarantees representation
for continuous multivariate functions, it does not necessarily encompass
discontinuous multivariate functions. For example, KART cannot be applied to
all multivariate operations as mentioned in (c) above, since such operations
may include division, which is not continuous.

But can the remarkable formula (1) in KART be \emph{extended} to discontinuous
multivariate functions? The answer is, fortunately, yes. A rigorous proof of
this extension can be found in \cite{Is1}. Thus, while the above statements are
accurate, they do not follow from the original KART; rather, the broader applicability
to discontinuous functions follows from the main result of \cite{Is1}.

Note that the latest generalization of KART not only encompasses all
multivariate functions (both continuous and discontinuous) but also replaces
the outer functions $\Phi _{q}$ with a single function $\Phi $, which
inherits continuity and boundedness properties of $f$. More precisely, the
following theorem holds.

\bigskip

\textbf{Theorem 1. }\cite{Isma} \textit{Assume$~n\geq 2$ is a given integer
and $\mathbb{I}=[0,1]$. There exist universal, continuous, one-variable
functions $\varphi _{q,p}$, $q=1,...,2n+1$, $p=1,...,n$, such that each $n$%
-variable function $f:\mathbb{I}^{n}\rightarrow \mathbb{R}$ (not necessarily
continuous) can be precisely expressed in the form}
\begin{equation*}
f(x_{1},...,x_{n})=\sum_{q=1}^{2n+1}\Phi (\sum_{p=1}^{n}\varphi
_{q,p}(x_{p})),\eqno(2)
\end{equation*}%
\textit{where $\Phi $ is a one-variable function depending on $f$. If $f$ is
continuous, then $\Phi $ can be chosen to be continuous as well, if $f$ is
discontinuous and bounded, then $\Phi $ is also discontinuous and bounded.
If $f$ is unbounded, then $\Phi $ is also unbounded.}

\bigskip

Note that $\varphi _{q,p}(x_{p})$ in (2) can be replaced by a family of
functions of simpler structure. Namely, instead of $\varphi _{q,p}(x_{p})$
one can take $\lambda _{p}\varphi (x_{p}+aq)$, where $\varphi $ is a single
fixed continuous function, $\lambda _{p}$ and $a$ are the explicitly given
real numbers (see \cite{Spr} and \cite{Isma}). Note also that Theorem 1 is
valid not only for the unit cube $[0,1]^{n}$ but for any closed cube $%
[a,b]^{n}$.

\bigskip

\textbf{Generalization to domains different from the unit cube:} There are a
large number of papers erroneously formulating KART for continuous
multivariate functions defined on bounded domains. The following formulation
is taken from \cite{Dong}:

(F1) \textquotedblleft Specifically, if $f$ is a continuous function on a
bounded domain $D\subset \mathbb{R}^{n}$, then there exist continuous
functions $\varphi _{ij}$ and $\psi _{i}$ such that
\begin{equation*}
f(x_{1},...,x_{n})=\sum_{i=1}^{2n+1}\psi _{i}(\sum_{j=1}^{n}\varphi
_{ij}(x_{j})),
\end{equation*}%
where $\varphi _{ij}:[0,1]\rightarrow \mathbb{R}$ and $\psi _{i}:\mathbb{%
R\rightarrow R}$."

The same formulation for $f$ on a bounded domain is given in the papers \cite%
{Ibrahum,Le,MLiu,Wang}.

However, the original KART can only be applied to continuous functions
defined on compact sets. The procedure for such a generalization is as
follows: Assume $K$ is a compact set in $\mathbb{R}^{n}$ and $S$ is a closed
cube of the form $[-a,a]^{n}$ containing $K$. By the Tietze extension
theorem \cite[Theorem 15.8]{Willard}, any $f\in C(K)$ can be continuously
extended to $S$. Denote this extension by $F.$ Since KART\ holds for $F$ on $%
S$, it also holds for $f$ on $K$.

The situation with a bounded domain $D\subset \mathbb{R}^{n}$ is completely
different. Although $D$ can embedded into a closed cube $S$, not every
function $f\in C(D)$ can be continuously extended to $S$. Consider, for
example, the function $f(x_{1},x_{2},x_{3})=x_{1}x_{2}x_{3}^{-1}$
on the open set $D=(0,1)^{3}$, which is a bounded domain. This function
cannot be continuously extended to any closed cube $S$ containing $D$.
Therefore, any extension of $f$ to $S$, denoted here by $F$, would
necessarily be discontinuous (and also unbounded). Clearly, we cannot apply
the original KART to $F$; hence we cannot apply it to $f$.

However, the statement in (F1) is valid. Although it does not follow from
the original KART, as erroneously indicated in the papers \cite%
{Dong,Ibrahum,Le,MLiu,Wang} and many other articles (which we do not mention
here), this statement easily follows from Theorem 1 above by applying
Theorem 1 to any (not necessarily continuous) extension of $f$ to a closed
cube $S$ containing $D$. We therefore formulate this as a corollary to
Theorem 1.

\bigskip

\textbf{Corollary 1. }\textit{Assume $D$ is a bounded domain in $\mathbb{R}%
^{n}$, $n\geq 2$. There exist universal continuous functions $\varphi _{q,p}:%
\mathbb{R}\rightarrow \mathbb{R}$, $q=1,...,2n+1$, $p=1,...,n$, such that
each $n$-variable function $f:D\rightarrow \mathbb{R}$ (not necessarily
continuous) can be precisely expressed in the form}
\begin{equation*}
f(x_{1},...,x_{n})=\sum_{q=1}^{2n+1}\Phi (\sum_{p=1}^{n}\varphi
_{q,p}(x_{p})),
\end{equation*}%
\textit{where $\Phi $ is a one-variable function depending on $f$.}

\bigskip

\bigskip

\begin{center}
{\large \textbf{Remark 2: Single-hidden-layer networks}}
\end{center}

Our second remark is about the universal approximation property of
single-hidden-layer neural networks. Such networks with $n$ units in the
hidden layer and input $\mathbf{x}=(x_{1},...,x_{d})$ compute a function of
the form
\begin{equation*}
\sum_{i=1}^{n}c_{i}\sigma (\mathbf{w}^{i}\mathbf{\cdot x}-\theta _{i}),\eqno%
(3)
\end{equation*}%
where the \textit{weights} $\mathbf{w}^{i}$ are vectors in $\mathbb{R}^{d}$,
the \textit{thresholds} $\theta _{i}$ and the \textit{coefficients} $c_{i}$
are real numbers and the \textit{activation function} $\sigma $ is a real
one-variable function. The Universal Approximation Theorem (UAT) plays an
essential role in neural network theory. This theorem says that
single-hidden-layer feedforward neural networks are capable of approximating
all continuous multivariate functions on compact subsets of the $d$%
-dimensional Euclidean space with arbitrary accuracy. That is, for a given
activation function $\sigma $, for any $\epsilon >0$, any compact subset $%
K\subset \mathbb{R}^{d}$ and any continuous function $f:K\rightarrow \mathbb{%
R}$ there exist $n(\epsilon )\in \mathbb{N}$, $\mathbf{w}^{i}\in \mathbb{R}%
^{d}$, $\theta _{i},c_{i}\in \mathbb{R}$ such that
\begin{equation*}
\max_{\mathbf{x}\in K}\left\vert f(\mathbf{x})-\sum_{i=1}^{n(\epsilon
)}c_{i}\sigma (\mathbf{w}^{i}\mathbf{\cdot x}-\theta _{i})\right\vert
<\epsilon .
\end{equation*}

The UAT holds for various classes of activation functions $\sigma $\ and
many methods have been developed to prove it. For a brief overview of some
of these methods, see \cite{Is2}. Notably, the most general result in this
area is due to Leshno, Lin, Pinkus and Schocken \cite{Leshno}, who proved
that single-hidden-layer neural networks with a continuous activation
function $\sigma $ have the universal approximation property if and only if $%
\sigma $ is not a polynomial.

Many papers discussing and reviewing UAT emphasize that the number of hidden
units $n(\epsilon )$ always depends on the approximation tolerance $\epsilon
$. That is, if we want to approximate continuous functions with arbitrarily
small precision, we necessarily need a large number of hidden neurons. For
example, the paper \cite{And}, published by IEEE, states: \textquotedblleft
Neural networks have the universal approximation property with respect to
continuous functions, i.e., the ability to approximate arbitrarily correctly
any continuous function, given that they have sufficiently many hidden
neurons. While this result holds in principle, in practice, the required
number of neurons may be excessively large".

There are hundreds of other papers in the literature stating that UAT
holds for single-hidden-layer networks, provided that the hidden layer
can contain arbitrarily many neurons (i.e., its width is unrestricted).

But should the number of units $n(\epsilon )$ really depend on $\epsilon $?
Actually, all known proofs for UAT are designed to validate this dependence.
However, is there any rigorous proof that UAT does not hold for shallow
networks with a fixed number of hidden units, implying that $n(\epsilon )$
must necessarily depend on $\epsilon$ ?

Such a mathematical proof can be found in \cite[Section 5]{Is}. It was
shown there that for $d>1$, and for any natural $N$, single-hidden-layer
networks with at most $N$ hidden units cannot approximate all continuous $d$%
-variable functions with arbitrary precision. Conversely, it was shown that
for $d=1$, the situation is drastically different. Specifically, in this
case, for certain activation functions and for any natural $N$,
single-hidden-layer networks with at most $N$ hidden units can approximate
all continuous univariate functions with arbitrary precision. Hence, in this
case, $n(\epsilon )$ does not depend on $\epsilon $. Specifically, the
following two theorems are valid.

\bigskip

\textbf{Theorem 2.} \cite{Is} \textit{For any positive number $\alpha $, there is a $%
C^{\infty }(\mathbb{R})$, almost monotone, sigmoidal activation function $%
\sigma _{\alpha }\colon \mathbb{R}\rightarrow \mathbb{R}$ satisfying the
following property: For any finite closed interval $[a,b]$ of $\mathbb{R}$
and any $f\in C[a,b]$ and $\varepsilon >0$ there exist three real numbers $%
c_{0}$, $c_{1}$ and $\theta $ for which}
\begin{equation*}
\left\vert f(x)-c_{1}\sigma _{\alpha }\left( \frac{\alpha }{b-a}x-\theta
\right) -c_{0}\right\vert <\varepsilon
\end{equation*}%
\textit{for all $x\in \lbrack a,b]$.}

\bigskip

\textbf{Theorem 3.} \cite{Is} \textit{Assume $d\geq 2$. For any continuous function $%
\sigma \colon \mathbb{R\rightarrow R}$, there is a $d$-variable continuous
function which cannot be approximated arbitrarily well by neural networks of
the form}
\begin{equation*}
\sum_{i=1}^{n}c_{i}\sigma (\mathbf{w}^{i}\cdot \mathbf{x}-\theta _{i}),
\end{equation*}%
\textit{where we vary over all $n\in \mathbb{N}$, $c_{i},\theta _{i}\in
\mathbb{R}$, $\mathbf{w}^{i}\in \mathbb{R}^{d}$, but the number of pairwise
independent vectors (weights) $\mathbf{w}^{i}$ in each network is uniformly
bounded by some positive integer $k$ (which is the same for all networks).}

\bigskip

Note that in Theorem 2, the parameters $c_{0}$, $c_{1}$ and $\theta $ can be
determined algorithmically for any Lipschitz continuous function $f$ (see
\cite{GI1}). It follows from Theorems 2 and 3 that if the number of hidden
neurons $n$ is fixed, then UAT\ holds if and only if the space dimension $%
d=1$.

For some nonsigmoidal activation functions Theorem 2 takes simpler form, in
which $c_{0}=0$ and $c_{1}=1$. More precisely, the following theorem is
valid.

\bigskip

\textbf{Theorem 4.} \textit{For any positive number $\alpha $, there is an
infinitely differentiable activation function $\sigma _{\alpha }\colon
\mathbb{R}\rightarrow \mathbb{R}$ such that for any finite closed interval $%
[a,b]$ of $\mathbb{R}$ and any $f\in C[a,b]$ and $\varepsilon >0$ there
exists a real number $\theta $ for which}
\begin{equation*}
\left\vert f(x)-\sigma _{\alpha }\left( \frac{\alpha }{b-a}x-\theta \right)
\right\vert <\varepsilon
\end{equation*}%
\textit{for all $x\in \lbrack a,b]$.}

\bigskip

\textbf{Proof.} We first consider the interval $[0,1]$. Let $\alpha $ be any
positive real number. Divide the interval $[\alpha ,+\infty )$ into the
segments $[\alpha ,2\alpha ],$ $[2\alpha ,3\alpha ],...$. Let $%
\{p_{n}(t)\}_{n=1}^{\infty }$ be the sequence of polynomials with rational
coefficients defined on $[0,1].$ Note that this sequence is dense in $C[0,1]$%
. We construct $\sigma _{\alpha }$ in two stages. In the first stage, we
define $\sigma _{\alpha }$ on the closed intervals $[(2m-1)\alpha ,2m\alpha
],$ $m=1,2,...$ as the function
\begin{equation*}
\sigma _{\alpha }(t)=p_{m}\left(\frac{t}{\alpha }-2m+1\right),\text{ }t\in
\lbrack (2m-1)\alpha ,2m\alpha ],
\end{equation*}%
or equivalently,
\begin{equation*}
\sigma _{\alpha }(\alpha t+(2m-1)\alpha )=p_{m}(t),\text{ }t\in \lbrack 0,1].%
\eqno(4)
\end{equation*}%
In the second stage, we extend $\sigma _{\alpha }$ to the intervals $%
(2m\alpha ,(2m+1)\alpha ),$ $m=1,2,...,$ and $(-\infty ,\alpha )$,
maintaining the $C^{\infty }$ property.

For any univariate function $h\in C[0,1]$ and any $\varepsilon >0$ there
exists a polynomial $p_{m}(t)$ with rational coefficients such that
\begin{equation*}
\left\vert h(t)-p_{m}(t)\right\vert <\varepsilon ,
\end{equation*}%
for all $t\in \lbrack 0,1].$ This together with (4) imply that for all $t\in
\lbrack 0,1]$ we have
\begin{equation*}
\left\vert h(t)-\sigma _{\alpha }(\alpha t-s)\right\vert <\varepsilon ,\eqno%
(5)
\end{equation*}%
where $s=(1-2m)\alpha .$

Using linear transformation it is not difficult to go from $[0,1]$ to any
finite closed interval $[a,b]$. Indeed, let $f\in C[a,b]$, $\sigma _{\alpha
} $ be constructed as above and $\varepsilon $ be an arbitrarily small
positive number. The transformed function $h(t)=f(a+(b-a)t)$ is well defined
on $[0,1]$ and we can apply the inequality (5). Now using the inverse
transformation $t=\frac{x-a}{b-a}$, we can write that
\begin{equation*}
\left\vert f(x)-\sigma _{\alpha }(wx-\theta )\right\vert <\varepsilon ,
\end{equation*}%
for all $x\in \lbrack a,b]$, where $w=\frac{\alpha }{b-a}$ and $\theta =%
\frac{\alpha a}{b-a}+s$.

\bigskip

\bigskip

\begin{center}
{\large \textbf{Remark 3: Deep neural networks}}
\end{center}

Regarding the universal approximation theorem for deep neural networks, it
is widely believed and emphasized in many studies that achieving a high
degree of accuracy in approximating multivariate functions requires large
networks with a sufficient number of hidden neurons. For example, the
well-known book \textit{Deep learning}\ by Goodfellow, Bengio and Courville
\cite{Good} states that \textquotedblleft there exists a neural network
large enough to achieve any degree of accuracy we desire, but the universal
approximation theorem does not say how large this network will be" (see
Chapter 6.4.1, Universal Approximation Properties and Depth in \cite{Good}).
Similar statements can be found in many other books and articles.

However, there exist neural networks with very few hidden neurons that can
approximate all continuous multivariate functions arbitrarily well.
Moreover, the number of hidden neurons required does not depend on the
desired approximation accuracy and can be determined precisely in advance.
For example, for $d$-variable continuous functions, this number can be as
small as $3d+2$ neurons distributed in two hidden layers. Furthermore, this
property holds even if all the weights are fixed. This means that for
certain activation functions, fixed weights and very few hidden neurons are
sufficient to achieve the universal approximation property. More precisely,
the following theorem holds:

\bigskip

\textbf{Theorem 5.} \cite{GI} \textit{One can algorithmically construct an
infinitely differentiable, almost monotone sigmoidal activation function $%
\sigma \colon \mathbb{R}\rightarrow \mathbb{R}$ satisfying the following
property: For any natural number $d\geq 2,$ any continuous function $f$ on
the unit cube $[0,1]^{d}$ and any $\varepsilon >0,$ there exist constants $%
e_{p}$, $c_{pq}$, $\theta _{pq}$ and $\zeta _{p}$ such that the inequality}
\begin{equation*}
\left\vert f(\mathbf{x})-\sum_{p=1}^{2d+2}e_{p}\sigma \left(
\sum_{q=1}^{d}c_{pq}\sigma (\mathbf{w}^{q}\cdot \mathbf{x}-\theta
_{pq})-\zeta _{p}\right) \right\vert <\varepsilon
\end{equation*}%
\textit{holds for all $\mathbf{x}=(x_{1},\ldots ,x_{d})\in \lbrack a,b]^{d}$%
. Here the weights $\mathbf{w}^{q}$, $q=1,\ldots ,d$, are fixed as follows:}
\begin{equation*}
\mathbf{w}^{1}=(1,0,\ldots ,0),\quad \mathbf{w}^{2}=(0,1,\ldots ,0),\quad
\ldots ,\quad \mathbf{w}^{d}=(0,0,\ldots ,1).
\end{equation*}%
\textit{In addition, all the coefficients $e_{p}$, except one, are equal.}

\bigskip

For detailed instructions on how to construct such activation functions $%
\sigma $ in practice and prove their universal approximation property,
consult \cite{GI} and \cite{Is}. It should be remarked that Maiorov and
Pinkus \cite{Mai} were the first to show that there exists an activation
function for which $9d+3$ neurons in the hidden layers of a two-hidden-layer
network are sufficient to approximate any continuous $d$-variable function
arbitrarily well. Their proposed activation function is sigmoidal, strictly
increasing, and analytic; however, they do not provide a feasible method for
its practical computation.

Recall that shallow networks with any fixed number of hidden neurons cannot
approximate $d$-variabe continuous functions if $d\geq 2$ (see Remark 2).

\bigskip

Note that for some nonsigmoidal activation functions, the number of hidden
neurons in the second hidden layer can be further reduced and coincide with
the number of outer terms in KART.

\bigskip

\textbf{Theorem 6.} \textit{There exists an infinitely differentiable
activation function $\sigma \colon \mathbb{R}\rightarrow \mathbb{R}$ with
the property: For any natural number $d\geq 2,$ any continuous function $f$
on the unit cube $[0,1]^{d}$ and any $\varepsilon >0,$ there exist constants
$\lambda _{q}$, $\theta _{p}$ and $\zeta $, for which the inequality}
\begin{equation*}
\left\vert f(\mathbf{x})-\sum_{p=1}^{2d+1}\sigma \left(
\sum_{q=1}^{d}\lambda _{q}\sigma (\mathbf{w}^{q}\cdot \mathbf{x}-\theta
_{p})-\zeta \right) \right\vert <\varepsilon
\end{equation*}%
\textit{holds for all $\mathbf{x}=(x_{1},\ldots ,x_{d})\in \lbrack 0,1]^{d}$%
. Here the weights $\mathbf{w}^{q}$, $q=1,\ldots ,d$, are fixed as follows:}
\begin{equation*}
\mathbf{w}^{1}=(1,0,\ldots ,0),\quad \mathbf{w}^{2}=(0,1,\ldots ,0),\quad
\ldots ,\quad \mathbf{w}^{d}=(0,0,\ldots ,1).
\end{equation*}

\bigskip

\textbf{Proof.} We use the following version of KART, attributed to Lorentz
\cite{Lor} and Sprecher \cite{Sp}: For the unit cube $[0,1]^{d},~d\geq 2,$
there exist constants $\lambda _{q}>0,$ $q=1,...,d,$ $\sum_{q=1}^{d}\lambda
_{q}=1,$ and nondecreasing continuous functions $h_{p}:[0,1]\rightarrow
\lbrack 0,1],$ $p=1,...,2d+1,$ such that every continuous function $%
f:[0,1]^{d}\rightarrow \mathbb{R}$ admits the representation

\begin{equation*}
f(x_{1},...x_{d})=\sum_{p=1}^{2d+1}g\left( \sum_{q=1}^{d}\lambda
_{q}h_{p}(x_{q})\right) \eqno(6)
\end{equation*}%
for some $g\in C[0,1]$ depending on $f.$

Assume we are given an arbitrary continuous function $f$ on $[0,1]^{d}$ and
any $\varepsilon >0$. Let $g$ be the outer function in (6). By Theorem 4,
there exists a smooth activation function $\sigma :=\sigma _{\alpha
}|_{\alpha =1}$ such that

\begin{equation*}
\left\vert g(t)-\sigma (t-\zeta )\right\vert <\frac{\varepsilon }{2(2d+1)},%
\eqno(7)
\end{equation*}%
for some $\zeta \in \mathbb{R}$ and all $t\in \lbrack 0,1].$

Taking into account (7) in (6), we obtain that

\begin{equation*}
\left\vert f(x_{1},...,x_{d})-\sum_{p=1}^{2d+1}\sigma \left(
\sum_{q=1}^{d}\lambda _{q}h_{p}(x_{q})-\zeta \right) \right\vert <\frac{%
\varepsilon }{2}\eqno(8)
\end{equation*}%
for all $(x_{1},...,x_{d})\in \lbrack 0,1]^{d}.$

Again, by Theorem 4, for each $p=1,2,...,2d+1,$ and any $\delta >0$ there
exists a constant $\theta _{p}$ such that

\begin{equation*}
\left\vert h_{p}(x_{q})-\sigma (x_{q}-\theta _{p})\right\vert <\delta ,\eqno%
(9)
\end{equation*}%
for all $x_{q}\in \lbrack 0,1].$ Since $\lambda _{q}>0,$ $q=1,...,d,$ $%
\sum_{q=1}^{d}\lambda _{q}=1,$ it follows from (9) that

\begin{equation*}
\left\vert \sum_{q=1}^{d}\lambda _{q}h_{p}(x_{q})-\sum_{q=1}^{d}\lambda
_{q}\sigma (x_{q}-\theta _{p})\right\vert <\delta ,\eqno(10)
\end{equation*}%
for all $(x_{1},...,x_{d})\in \lbrack 0,1]^{d}.$

Now since the function $\sigma (t-\zeta )$ is uniformly continuous on every
closed interval, we can choose $\delta $ to be sufficiently small, and from
(10), we obtain that

\begin{equation*}
\left\vert \sum_{p=1}^{2d+1}\sigma \left( \sum_{q=1}^{d}\lambda
_{q}h_{p}(x_{q})-\zeta \right) -\sum_{p=1}^{2d+1}\sigma \left(
\sum_{q=1}^{d}\lambda _{q}\sigma (x_{q}-\theta _{p})-\zeta \right)
\right\vert <\frac{\varepsilon }{2}.\eqno(11)
\end{equation*}%
It follows from (8) and (11) that

\begin{equation*}
\left\vert f(\mathbf{x})-\sum_{p=1}^{2d+1}\sigma \left(
\sum_{q=1}^{d}\lambda _{q}\sigma (\mathbf{w}^{q}\cdot \mathbf{x-}\theta
_{p})-\zeta \right) \right\vert <\varepsilon ,
\end{equation*}%
where $\mathbf{w}^{q}$ is the $q$-th coordinate vector. The theorem has been
proved.

\bigskip

Note that Theorems 5 and 6 hold for any compact subset of $\mathbb{R}^{d}$,
as KART is valid not only for the unit cube but for compact sets in general.

\bigskip

\bigskip

\begin{center}
{\large \textbf{Remark 4: Minimum number of neurons for universal
approximation}}
\end{center}

In the theory of deep neural networks, the problem of determining the
minimum width that guarantees the universality of networks has recently
gained considerable attention from researchers. See, e.g., \cite{Cai,Park},
for an extensive collection of references and comparisons of various
results. These results primarily focus on a critical threshold for the width
that allows deep neural networks to be universal approximators. However,
none of these results address the problem of finding the minimum number of
neurons required to ensure universality.

In machine learning, the question often arises: how many neurons should be
in a fully connected layer of a neural network to solve a given problem
correctly? Investigating this question, \cite{Ber}, in particular, writes
that for fully connected networks, KART provides an answer: specifically,
KART shows that for a two-hidden-layer network, the minimum number of
neurons required in the second layer is $2N_{in}+1$, where $N_{in}$ is the
input dimension, $N_{out}=1$ is the output dimension. However, KART proposes
a specific neural network structure that differs from traditional multilayer
feedforward networks. Fortunately, the question of whether this holds true
for conventional MLPs (multilayer perceptrons) is answered affirmatively, as
shown in Theorem 6. Moreover, the activation function in this theorem is
ultimately smooth, which contrasts with the activation functions in
KART-induced networks, where the inner activation functions $\varphi _{q,p}$
can be at most from the Lipschitz class $Lip(1)$ (see \cite[Chapter 4]{Is}
for discussions and references).

Theorem 4 above directly implies that \textit{the minimal number of hidden
neurons needed for neural networks to approximate any continuous univariate
function is exactly $1$.} The situation with continuous multivariate
functions is more complex, but based on Sternfeld's results on the minimal
number of terms in KART, we conjecture that Theorem 6, in particular,
establishes the minimum number of neurons for universal approximation. For
an overview of Sternfeld's results concerning KART, see \cite[Chapter 1]{Kha}%
.

\smallskip

\textbf{Conjecture.} \textit{For a deep network with $d$ inputs ($d>1$) and
a single output, the minimum number of hidden neurons required to ensure
universality is $3d+1$. More precisely, neural networks with fewer than $%
3d+1 $ hidden neurons cannot approximate $d$-variable continuous functions
with arbitrary precision.}

\medskip

\textbf{Comparison with related results.} The problem of determining the
minimal width $w_{\min }$ required for universal approximation has been the
subject of several studies. Let $\mathcal{K}\subset \mathbb{R}^{d_{x}}$ be
compact, and consider continuous target functions with values in $\mathbb{R}%
^{d_{y}}$. Some representative results include the following.

Hanin and Sellke~\cite{Hanin} showed that for ReLU networks approximating
functions in $C(\mathcal{K},\mathbb{R})$, the minimal width is $w_{\min
}=d_{x}+1$. Park, Yun, Lee, and Shin~\cite{Park} extended this to
vector-valued functions in $C(\mathcal{K},\mathbb{R}^{d_{y}})$, proving that
ReLU+STEP networks require $w_{\min }=\max (d_{x}+1,d_{y})$. More generally,
Cai~\cite{Cai} established that for arbitrary activation functions, $w_{\min
}\geq \max (d_{x},d_{y})$, and for ReLU+FLOOR networks one has $w_{\min
}=\max (d_{x},d_{y},2)$. Gripenberg~\cite{Gripenberg} showed that if the activation
function is twice differentiable at some point with nonzero second
derivative, then $w_{\min }\leq d_{x}+d_{y}+2$. Johnson~\cite{Johnson} proved
that for functions in $C(\mathcal{K},\mathbb{R}^{d_{y}})$, when the
activation function is uniformly continuous and can be approximated by a
sequence of one-to-one functions, one has $w_{\min }\geq d_{x}+1$.

All of the above results require networks of arbitrarily large depth. By contrast, within the framework of this paper, Theorems~6 shows that universal approximation for functions in $C(\mathcal{K},\mathbb{R})$ can already be achieved by networks of \emph{fixed depth} (depth $=3$) and width $2d_{x}+1$. This establishes the upper bound $w_{\min}\leq 2d_{x}+1$ in the fixed-depth setting. In particular, for $d_{x}=1$, Theorem~4 shows that $w_{\min}=1$. Moreover, we conjecture that the exact minimal width for $d_{x}>1$ is $w_{\min}=2d_{x}+1$. We also note that the earlier result of Maiorov and Pinkus~\cite{Mai} required width $6d_x+3$ at depth $3$ for functions in $C(\mathcal{K},\mathbb{R})$, showing that our construction improves this bound.

\medskip

\begin{table}[ht]
\caption{Summary of minimum width of neural networks that have the UAP}
\begin{center}
\begin{tabular}{p{2.8cm} p{5cm} p{4cm} p{2.2cm}}
\multicolumn{1}{c}{\bf Functions} & \multicolumn{1}{c}{\bf Activation} & \multicolumn{1}{c}{\bf Minimum width} & \multicolumn{1}{c}{\bf References}\\
\hline
$C(\mathcal{K},\mathbb{R})$ & ReLU (arbitrarily large depth) & $w_{\min} = d_x+1$ & Hanin and Sellke \cite{Hanin} \\
$C(\mathcal{K},\mathbb{R}^{d_y})$ & ReLU+STEP (arbitrarily large depth) & $w_{\min} = \max(d_x+1,d_y)$ & Park et al. \cite{Park} \\
$C(\mathcal{K},\mathbb{R}^{d_y})$ & Arbitrary (arbitrarily large depth) & $w_{\min} \ge \max(d_x,d_y)$ & Cai \cite{Cai} \\
$C(\mathcal{K},\mathbb{R}^{d_y})$ & ReLU+FLOOR (arbitrarily large depth) & $w_{\min} = \max(d_x,d_y,2)$ & Cai \cite{Cai} \\
$C(\mathcal{K},\mathbb{R}^{d_y})$ & $C^2$ at some $z$ with $\sigma''(z)\neq 0$ (arbitrarily large depth) & $w_{\min} \le d_x+d_y+2$ & Gripenberg \cite{Gripenberg} \\
$C([0,1]^{d_x},\mathbb{R}^{d_y})$ & Uniformly continuous, approximable by one-to-one functions (arbitrarily large depth) & $w_{\min} \ge d_x+1$ & Johnson \cite{Johnson} \\
\hline
$C(\mathcal{K},\mathbb{R})$ & Specific analytic sigmoidal (depth $=3$) & $w_{\min} \le 6d_x+3$ & Maiorov and Pinkus \cite{Mai} \\
$C(\mathcal{K},\mathbb{R})$ & Specific $C^\infty$ (depth $=3$) & $w_{\min} \le 2d_x+1$ & {\bf Ours} (Thm.~6) \\
$C([a,b],\mathbb{R})$ & Specific $C^\infty$ (depth $=2$) & $w_{\min} = 1$ & {\bf Ours} (Thm.~4) \\
\hline
\end{tabular}
\end{center}
\smallskip
\noindent\textit{Note:} The minimal width bounds listed in the table for networks of fixed depth (2 or 3) also apply to deeper networks, as additional layers cannot increase $w_{\min}$ and may only reduce it.
\label{tab:min_width}
\end{table}

\bigskip

\begin{center}
{\large \textbf{Conclusion}}
\end{center}

In this note, we have addressed the growing misinterpretations related to
the Kolmogorov-Arnold Representation Theorem (KART) and the Universal
Approximation Theorem (UAT) in the context of neural network approximation.
We highlighted the importance of correctly understanding the scope and
limitations of KART --- particularly in distinguishing its actual
implications for continuous functions from the widespread misconceptions
that arise when treating the theorem as if it were originally proven for
arbitrary multivariate functions or for continuous functions defined on
arbitrary bounded domains, which is not the case. We also discussed UAT,
focusing on the number of hidden neurons required for approximating
continuous functions with arbitrary precision, and we provided a clearer
understanding on how this number may depend on the dimension of the input
layer. By addressing these common and frequent misconceptions, which have
become increasingly prevalent in the current literature, we aim to encourage
a more accurate and rigorous use of KART, UAT, and their refinements and
generalizations in future research.

We also addressed a related but distinct question: what is the minimum
number of neurons required to ensure the universality of deep neural
networks? While much of the recent literature concentrates on the minimal
width necessary for universal approximation, we examined the problem from
the perspective of neuron count. Building on KART and the structural
properties it suggests, we showed --- through Theorem 6 --- that the same
number of neurons required for exact representation of functions in
KART-based networks also applies to standard multilayer perceptrons in the
context of universal approximation. Importantly, this holds even when the
activation functions are smooth, unlike in KART, where only Lipschitz
continuous activations are used.

For univariate functions, Theorem 4 confirms that only one hidden neuron is
needed to achieve universality. In the multivariate case, drawing from
Sternfeld's analysis of KART, we proposed a conjecture: a network with fewer
than $3d+1$ hidden neurons cannot approximate arbitrary continuous functions
of $d$ variables. This observation suggests a natural lower bound that
complements existing results on width and offers a new understanding of the
architecture of universal approximators.

\medskip

\noindent\textbf{Practical relevance.} Although our results are primarily theoretical, they also carry practical significance. First, the constructive aspects of Theorems~2, 4, 5, and 6 provide explicit neuron counts and fixed-weight constructions that can guide future implementations. Second, the demonstrated generality (including discontinuous and unbounded functions) highlights the expressive power of KART-based networks, which may motivate new variants and extensions of KANs. Finally, we note that even the original KART required decades before it found practical realization in neural architectures. In the same spirit, our results, especially the generalization of KART to all possible functions, should be viewed as establishing a rigorous theoretical foundation on which future practical advances can be built, rather than as immediately implementable methods.

\bigskip

\begin{center}
{\large \textbf{Acknowledgements}}
\end{center}

The author is grateful to the reviewers for their insightful comments and constructive suggestions, which have improved the clarity and quality of the manuscript.


\bigskip

\begin{thebibliography}{99}
\bibitem{And} P. Andras, High-dimensional function approximation with neural
networks for large volumes of data, \textit{IEEE Trans. Neural Netw. Learn.
Syst.} \textbf{29} (2018), no. 2, 500--508.

\bibitem{Ber} O. I. Berngardt, Minimum number of neurons in fully connected
layers of a given neural network (the first approximation), arXiv preprint,
arXiv:2405.14147, 2024.

\bibitem{Boz} Z. Bozorgasl, H. Chen, Wav-KAN: Wavelet Kolmogorov-Arnold
networks, arXiv preprint, arXiv:2405.12832, 2024.

\bibitem{Bresson} R. Bresson, G. Nikolentzos, G. Panagopoulos, M.
Chatzianastasis, J. Pang, M. Vazirgiannis, KAGNNs: Kolmogorov-Arnold
networks meet graph learning, arXiv preprint, arXiv:2406.18380, 2024.

\bibitem{Cai} Y. Cai, Achieve the minimum width of neural networks for
universal approximation, in The Eleventh International Conference on
Learning Representations, 2023.

\bibitem{Carlo} G. De Carlo, A. Mastropietro, A. Anagnostopoulos,
Kolmogorov-Arnold graph neural networks, arXiv preprint, arXiv:2406.18354,
2024.

\bibitem{Dong} C. Dong, L. Zheng, W. Chen, Kolmogorov-Arnold networks (KAN)
for time series classification and robust analysis. In: Sheng, Q.Z., et al.
Advanced Data Mining and Applications. ADMA 2024. Lecture Notes in Computer
Science, vol 15390 (2025). Springer, Singapore.

\bibitem{Good} I. Goodfellow, Y. Bengio, A. Courville, \textit{Deep learning}%
, MIT Press, Cambridge, MA, 2016.

\bibitem{Gripenberg} G. Gripenberg, Approximation by neural networks with a bounded
number of nodes at each level, \textit{J. Approx. Theory} \textbf{122} (2003),
no. 2, 260-266.

\bibitem{GI1} N. J. Guliyev, V. E. Ismailov, A single hidden layer
feedforward network with only one neuron in the hidden layer can approximate
any univariate function, \textit{Neural Comput.} \textbf{28} (2016), no. 7,
1289--1304.

\bibitem{GI} N. J. Guliyev, V. E. Ismailov, Approximation capability of two
hidden layer feedforward neural networks with fixed weights, \textit{%
Neurocomputing} \textbf{316} (2018), 262--269.

\bibitem{Hanin} B. Hanin and M. Sellke, Approximating continuous functions by
ReLU nets of minimal width, arXiv preprint, arXiv:1710.11278v2, 2018.

\bibitem{Ibrahum} A. D. M. Ibrahum, Z. Shang, J.-E. Hong, How resilient are
Kolmogorov--Arnold networks in classification tasks? A robustness
investigation, \textit{Appl. Sci.} \textbf{14} (2024), 10173.

\bibitem{Is} V. E. Ismailov, \textit{Ridge functions and applications in
neural networks}, Mathematical Surveys and Monographs 263, American
Mathematical Society, 2021, 186 pp.

\bibitem{Is1} V. E. Ismailov, A three layer neural network can represent any
multivariate function, \textit{J. Math. Anal. Appl.} \textbf{523} (2023),
no. 1, Paper No. 127096, 8 pp.

\bibitem{Is2} V. E. Ismailov, Approximation error of single hidden layer
neural networks with fixed weights, \textit{Inform. Processing Lett.}
\textbf{185} (2024), Paper No. 106467

\bibitem{Isma} A. Ismayilova, V. E. Ismailov, On the Kolmogorov neural
networks, \textit{Neural Netw.} \textbf{176} (2024), Paper No. 106333.

\bibitem{Johnson} J. Johnson, Deep, skinny neural networks are not universal
approximators, In: International Conference on Learning Representations, 2019.

\bibitem{Kha} S. Ya. Khavinson, \textit{Best approximation by linear
superpositions (approximate nomography),} Translated from the Russian
manuscript by D. Khavinson. Translations of Mathematical Monographs, 159.
American Mathematical Society, Providence, RI, 1997, 175 pp.

\bibitem{Kia} M. Kiamari, M. Kiamari, B. Krishnamachari, GKAN: Graph
Kolmogorov-Arnold networks, arXiv preprint, arXiv:2406.06470, 2024.

\bibitem{Kol} A. N. Kolmogorov, On the representation of continuous
functions of many variables by superposition of continuous functions of one
variable and addition. (Russian), \textit{Dokl. Akad. Nauk SSSR} \textbf{114}
(1957), 953--956.

\bibitem{Le} N. Le, A. P. Ngo, H. T. Nguyen, Kolmogorov-Arnold networks for
supervised learning tasks in smart grids, 56th North American Power
Symposium (NAPS), El Paso, TX, USA, 2024, 1-6.

\bibitem{Leshno} M. Leshno, V. Ya. Lin, A. Pinkus, S. Schocken, Multilayer
feedforward networks with a non-polynomial activation function can
approximate any function, \textit{Neural Netw.} \textbf{6} (1993), 861--867.

\bibitem{MLiu} M. Liu, D. Geibler, D. Nshimyimana, S. Bian, B. Zhou, P.
Lukowicz, Initial investigation of Kolmogorov-Arnold networks (KANs) as
feature extractors for IMU based human activity recognition, Companion of
the 2024 on ACM International Joint Conference on Pervasive and Ubiquitous
Computing, 2024, 500-506.

\bibitem{Liu} Z. Liu, Y. Wang, S. Vaidya, F. Ruehle, J. Halverson, M. Solja\v{c}i\'{c}, T. Y. Hou, M. Tegmark, KAN: Kolmogorov-Arnold networks, arXiv
preprint, arXiv:2404.19756, 2024.

\bibitem{Liu2} Z. Liu, P. Ma, Y. Wang, W. Matusik, M. Tegmark, KAN 2.0:
Kolmogorov-Arnold networks meet science, arXiv preprint, arXiv:2408.10205,
2024.

\bibitem{Lor} G. G. Lorentz, Metric entropy, widths, and superpositions of
functions. \textit{Amer. Math. Monthly} \textbf{69} (1962), 469--485.

\bibitem{Mai} V. Maiorov, A. Pinkus, Lower bounds for approximation by MLP
neural networks, \textit{Neurocomputing} \textbf{25} (1999), 81--91.

\bibitem{Park} S. Park, C. Yun, J. Lee, J. Shin, Minimum width for universal
approximation, In International Conference on Learning Representations, 2021.

\bibitem{Peng} Y. Peng, Y. Wang, F. Hu, M. He, Z. Mao, X. Huang, J. Ding,
Predictive modeling of flexible EHD pumps using Kolmogorov--Arnold Networks,
\textit{Biomimetic Intell. Robot.} \textbf{4} (2024), Issue. 4, Article No.
100184.

\bibitem{Sp} D. A. Sprecher, On the structure of continuous functions of
several variables. \textit{Trans. Amer. Math. Soc.} \textbf{115} (1965),
340--355.

\bibitem{Spr} D. A. Sprecher, A numerical implementation of Kolmogorov's
superpositions, \textit{Neural Netw.} \textbf{9} (1996), 765--772.

\bibitem{Wang} Z. Wang, A. Zainal, M. M. Siraj, F. A. Ghaleb, X. Hao, S.
Han, An intrusion detection model based on Convolutional Kolmogorov-Arnold
Networks, \textit{Sci. Rep.} \textbf{15} (2025), Article No. 1917.

\bibitem{Willard} S. Willard, \textit{General topology}, Addison-Wesley
Publishing Co., Reading, Mass.-London-Don Mills, Ont., 1970.


\end{thebibliography}
\end{document}